\documentclass[11pt,letterpaper,fleqn]{article}
\usepackage{naaclhlt2015}
\usepackage{times}
\usepackage{latexsym}

\usepackage{url}
\usepackage{array}
\usepackage{graphicx}
\usepackage{wrapfig}
\usepackage{subfigure}
\usepackage{color,xcolor}
\usepackage{amsmath}
\usepackage{multirow}
\usepackage{amssymb}
\usepackage{pifont}

\newcommand{\nascomment}[1]{\textcolor{blue}{\textsc{[#1 --NAS]}}}
\newcommand{\flcomment}[1]{\textcolor{magenta}{\textsc{[#1 --FL]}}}
\newcommand{\jmf}[1]{\textcolor{green}{[#1 --JMF]}}
\newcommand{\smtcomment}[1]{\textcolor{orange}{[#1 --SMT]}}

\renewcommand{\nascomment}[1]{}
\renewcommand{\flcomment}[1]{}
\renewcommand{\jmf}[1]{}
\renewcommand{\smtcomment}[1]{}

\title{Toward Abstractive Summarization Using Semantic Representations}

\author{Fei Liu \quad Jeffrey Flanigan \quad Sam Thomson \quad Norman Sadeh \quad Noah A. Smith \\
	    School of Computer Science\\
	    Carnegie Mellon University\\
	    Pittsburgh, PA 15213, USA\\
	    {\tt \{feiliu, jflanigan, sthomson, sadeh, nasmith\}@cs.cmu.edu}
}

\date{}

\begin{document}
\maketitle

\begin{abstract}

We present a novel abstractive summarization framework 
that draws on the recent development of a treebank for the Abstract Meaning Representation (AMR).
In this framework, the source text is parsed to a set of AMR graphs, the
graphs are transformed into a summary graph, and then text is generated from the summary graph.
We focus on the graph-to-graph transformation that reduces the source
semantic graph into a summary graph, making use of an existing AMR
parser and assuming the eventual availability of  an AMR-to-text
generator.
The framework is data-driven, trainable, and not specifically
designed for a particular domain.
Experiments on gold-standard AMR annotations and system parses show promising results.
Code is available at: {\fontfamily{lmtt}\selectfont https://github.com/summarization}
\smtcomment{Something about releasing your code?}
\flcomment{best place to host code repository?}

\end{abstract}

\section{Introduction}
\label{sec:intro}

Abstractive summarization is an elusive technological capability in
which textual summaries of content are generated \emph{de novo}.
Demand is on the rise for high-quality summaries not just for lengthy texts
(e.g., books; Bamman and Smith, 2013)
and texts known to be prohibitively difficult for people to understand
(e.g.,
website privacy policies; Sadeh et al., 2013),
\nocite{Sadeh:2013,Bamman:2013} but also for non-textual media (e.g., videos and
image collections; Kim et al., 2014; Kuznetsova et al., 2014; Zhao and
Xing, 2014), \nocite{Kuznetsova:2014,Kim:2014,Zhao:2014} where
extractive and compressive summarization techniques simply do not
suffice.
We believe that the challenge of abstractive summarization
deserves renewed attention and propose that recent developments in
semantic analysis have an important role to play.

\begin{figure}[t]
\begin{center}
\includegraphics [width=3.2in] {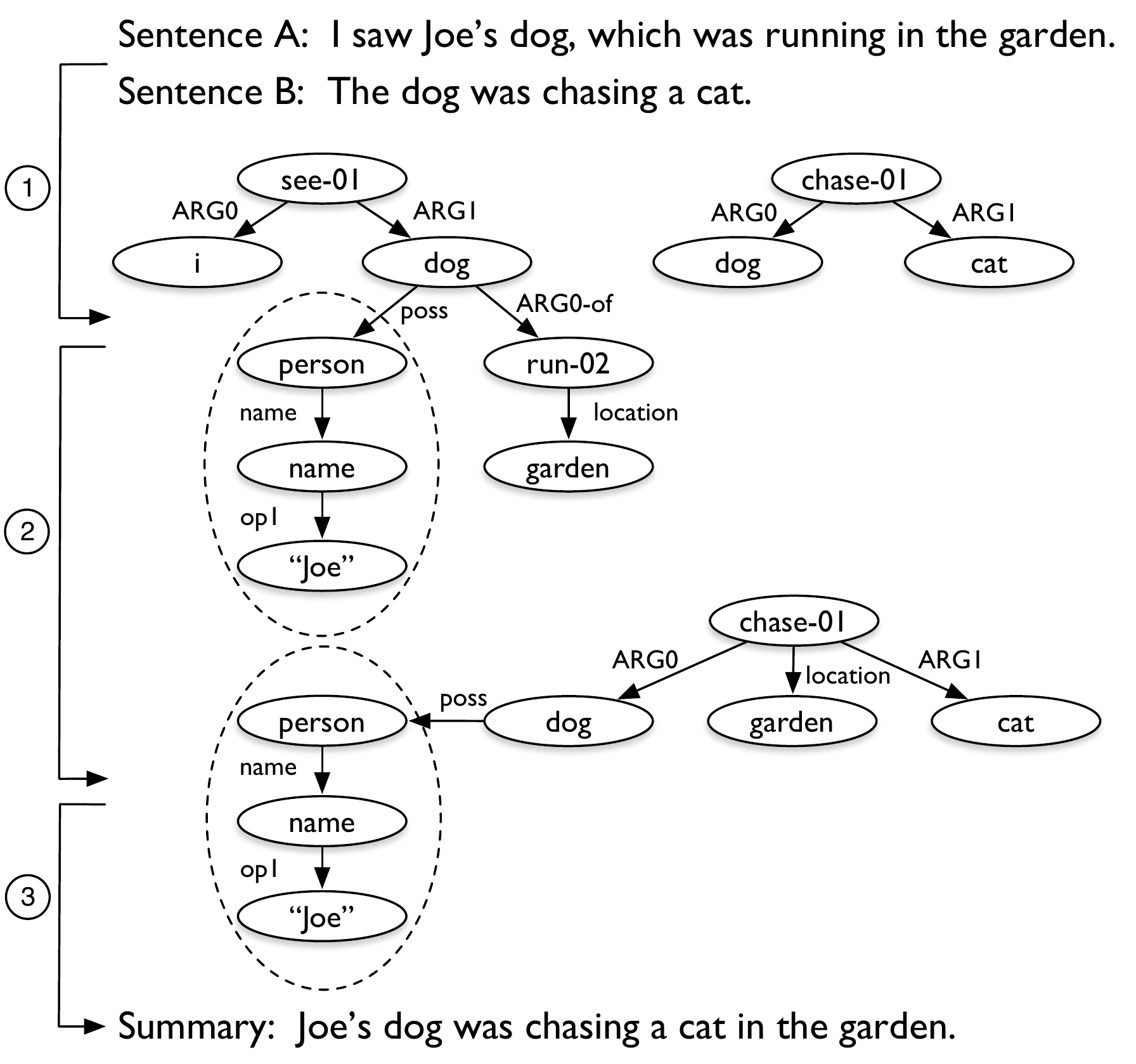}
\caption{A toy example. Sentences are parsed into individual AMR graphs in step 1; step 2 conducts graph transformation that produces a single summary AMR graph; text is generated from the summary graph in step 3.}
\label{fig:amr_graph}
\end{center}
\end{figure}

We conduct the first study exploring the feasibility of
an abstractive summarization system based on transformations of
semantic representations such as the Abstract Meaning
Representation (AMR; Banarescu et al., 2013). \nocite{Banarescu:2013}
Example sentences and their AMR graphs are shown in
Fig.~\ref{fig:amr_graph}.
AMR has much in common with earlier formalisms
\cite{Kasper:1989,dorr_thematic_1998}; today an annotated corpus
comprised of over 20,000 
AMR-analyzed English sentences \cite{Knight:2014}
and an automatic AMR parser (JAMR; Flanigan et al., 2014) \nocite{Flanigan:2014} are available.

%

In our framework, summarization consists of three steps illustrated in
Fig.~\ref{fig:amr_graph}:  (1) parsing
the input sentences to individual AMR graphs, (2) combining and transforming those graphs into a single summary
AMR graph, and (3) generating text from the summary graph.  This
paper focuses on step 2, treating it as a structured prediction problem. We assume text documents as
input\footnote{In principle, the framework could be applied to other
  inputs, such as image collections, if AMR parsers became available for
  them.} and use JAMR for step 1.  We use a simple method to read a
bag of words off the summary graph, allowing evaluation with ROUGE-1, 
 and leave full text
generation from AMR (step 3) to future work.

The graph summarizer, described in \S\ref{sec:strpred}, first
merges
 AMR graphs for each input sentence through a \emph{concept merging} step, in which coreferent nodes of the graphs are merged; 
a \emph{sentence conjunction} step, which connects the root of each sentence's AMR graph to a dummy ``ROOT'' node;
and an optional \emph{graph expansion} step, where additional edges are
added to create a fully dense graph on the sentence-level.
These steps result in a single connected \emph{source graph}.
A subset of the nodes and arcs from the source graph are then selected
for inclusion in the \emph{summary graph}.  Ideally this is a condensed
representation of the most salient semantic content from the source.

We briefly review AMR and JAMR (\S\ref{sec:amr}), then present the dataset used in this
paper (\S\ref{sec:data}).  The main algorithm is presented in
\S\ref{sec:strpred}, and we discuss our simple generation step in \S\ref{sec:nlg}.
Our experiments (\S\ref{sec:experiments}) measure the intrinsic quality of the graph
transformation algorithm as well as the quality of the terms selected
for the summary (using ROUGE-1).  We explore variations on the
transformation and the learning algorithm, and show oracle upper
bounds of various kinds.

\section{Background: Abstract Meaning Representation and JAMR}
\label{sec:amr}

AMR provides a whole-sentence semantic representation, represented as
a rooted, directed, acyclic graph (Fig.~\ref{fig:amr_graph}). 
Nodes of an AMR graph are labeled with \emph{concepts}, and edges are labeled with \emph{relations}.
Concepts can be English words (``dog''), PropBank event predicates (``chase-01,'' ``run-02''), or special keywords (``person'').
For example, ``chase-01'' represents a PropBank roleset that corresponds to the first sense of ``chase''.
According to ~\newcite{Banarescu:2013}, AMR uses approximately 100 relations.
The rolesets and core semantic relations (e.g., ARG0 to ARG5) are adopted from the PropBank annotations in OntoNotes~\cite{Hovy:2006}.
Other semantic relations include ``location,'' ``mode,'' ``name,''
``time,'' and ``topic.''
The AMR
guidelines\footnote{\url{http://www.isi.edu/~ulf/amr/help/amr-guidelines.pdf}}
provide more detailed descriptions.  \newcite{Banarescu:2013} describe
AMR Bank, a 20,341-sentence corpus annotated with AMR by experts.



Step 1 of our framework converts input document sentences into AMR graphs.
We use a statistical semantic parser, JAMR~\cite{Flanigan:2014}, which
was trained on AMR Bank.
JAMR's current performance on our test dataset  is 63\% $F$-score.\footnote{AMR parse quality is
  evaluated using smatch \cite{Cai:2013}, which measures the
  accuracy of concept and relation predictions.  JAMR was trained on
  the in-domain training portion\nascomment{I don't understand why the word ``proxy'' is here}\flcomment{slightly reworded.} of LDC2014T12 for our experiments.}
We will analyze the effect of AMR parsing errors by comparing JAMR
output with gold-standard annotations of input sentences in the experiments (\S\ref{sec:experiments}).

In addition to predicting AMR graphs for each sentence, JAMR provides
alignments between spans of words in the source sentence and fragments
of its predicted graph.  For example, 
a graph fragment headed by ``date-entity'' could be aligned to the tokens
``April 8, 2002.''  We use these alignments in our simple text
generation module (step 3; \S\ref{sec:nlg}).





\section{Dataset}
\label{sec:data}

To build and evaluate our framework, we require a dataset that
includes inputs and summaries, each with gold-standard AMR
annotations.\footnote{Traditional multi-document summarization
  datasets, such as the ones used in DUC and TAC competitions, do not
  have gold-standard AMR annotations.} 
 This allows us to use a statistical model for step 2
(graph summarization) and to separate its errors from those in step 1
(AMR parsing), which is important in determining whether this approach
is worth further investment.  

Fortunately, the ``proxy report''
section of the AMR Bank \cite{Knight:2014} suits our needs.
A proxy report is created by annotators based on a single newswire article,
selected from the English Gigaword corpus.
The report header contains metadata about date, country, topic, and a short summary. 
The report body is generated by editing or rewriting the content of the newswire  article to approximate the style of an analyst report.
Hence this is a single document summarization task.
All sentences are paired with gold-standard AMR annotations.
Table~\ref{tab:data} provides an overview of our dataset.

\begin{table}[t]
\setlength{\tabcolsep}{2.5pt}
\begin{center}\small
\begin{tabular}{|c|r|rr|rrr|}
\cline{2-7}
\multicolumn{1}{c|}{}
& \multirow{2}{*}{\bf \# Docs.} & \multicolumn{2}{c|}{\bf Ave.~\# Sents.} & \multicolumn{3}{c|}{\textbf{Source Graph}}\\
\multicolumn{1}{c|}{} &  & Summ. & Doc. & Nodes & Edges & Expand\\
\cline{2-7}
\hline
Train & 298 & 1.5 & 17.5 & 127 & 188 & 2,670 \\
Dev. & 35 & 1.4 & 19.2 & 143 & 220 & 3,203 \\
Test & 33 & 1.4 & 20.5 & 162 & 255 & 4,002\\
\hline
\end{tabular}
\caption{Statistics of our dataset.  ``Expand'' shows the number of
  edges after performing graph expansion. The numbers are averaged 
  across all documents in the split. We use the official split, 
  dropping one training document for which no summary sentences were annotated.
\label{tab:data}}
\end{center}
\vspace{0.05in}
\end{table}



\section{Graph Summarization}
\label{sec:strpred}

Given AMR graphs for all of the sentences in the input (step 1), graph
summarization transforms them into a single summary AMR graph (step 2).
This is accomplished in two stages:
source graph construction (\S\ref{subsec:sourcegraph}); and
subgraph prediction (\S\ref{subsec:subgraphpred}).


\subsection{Source Graph Construction}
\label{subsec:sourcegraph}

The ``source graph'' is a single graph constructed using the
individual sentences' AMR graphs by merging identical concepts.
In the AMR formalism, an entity or event is canonicalized and
represented by a single graph fragment, regardless of how many times
it is referred to in the sentence.  This principle can be extended to
multiple sentences, ideally resulting in a source graph with no
redundancy.  Because repeated mentions of a concept in the input can
signal its importance, we will later encode the frequency of mentions
as a feature used in subgraph prediction.

\begin{figure}[t]
\begin{center}
\includegraphics [width=3in] {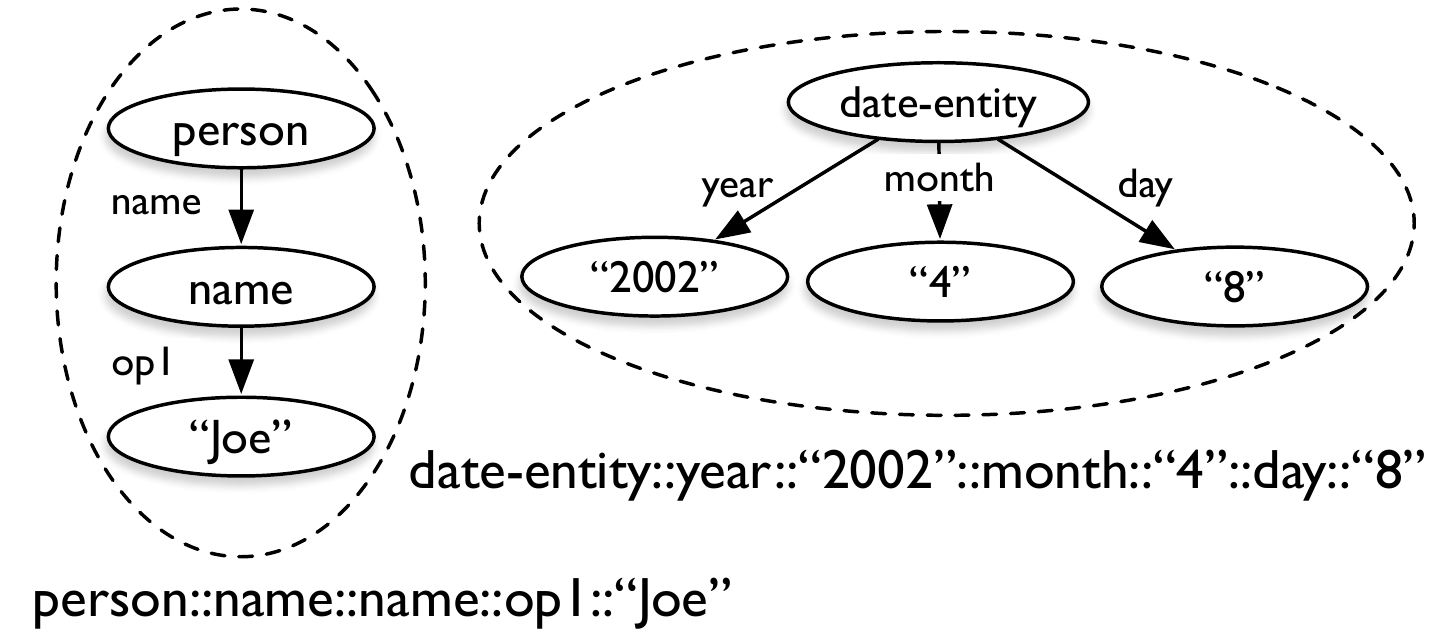}
\caption{Graph fragments are collapsed into a single concept and assigned a new concept label. 
\smtcomment{might be nice to draw an oval around the new concatenated labels, to emphasize that they're now concepts.}\flcomment{switched to oval for representing  collapsed graph fragments.. hope it has a similar effect.}
}
\label{fig:collapsing_example}
\end{center}
\end{figure}

Concept merging involves collapsing certain graph fragments into a single concept, then merging all concepts that have the same label.
We collapse the graph fragments that are headed by either a date-entity (``date-entity'') or a named entity (``name''), if the fragment is a flat structure.
A collapsed named entity is further combined with its parent (e.g., ``person'') into one concept node if it is the only child of the parent.
Two such graph fragments are illustrated in Fig.~\ref{fig:collapsing_example}.
We choose named and date entity concepts since they appear frequently, but most often refer to different entities (e.g., ``April 8, 2002'' vs. ``Nov. 17'').
No further collapsing is done.
A collapsed graph fragment is assigned a new label by concatenating the consisting concept and edge labels.
Each fragment that is collapsed into a new concept node can then only be merged with other identical fragments.
This process won't recognize coreferent concepts like ``Barack Obama'' = ``Obama'' and ``say-01'' = ``report-01,'' but future work may incorporate both entity coreference resolution and event coreference resolution, as concept nodes can represent either.

Due to the concept merging step, a pair of concepts may now have multiple labeled edges between them.
We merge all such edges between a given pair of concepts into a single unlabeled edge. 
We remember the two most common 
labels in such a group, which are used in the edge ``Label" feature (Table~\ref{tab:features}).


\begin{figure}[t]
\begin{center}
\includegraphics [width=3.2in] {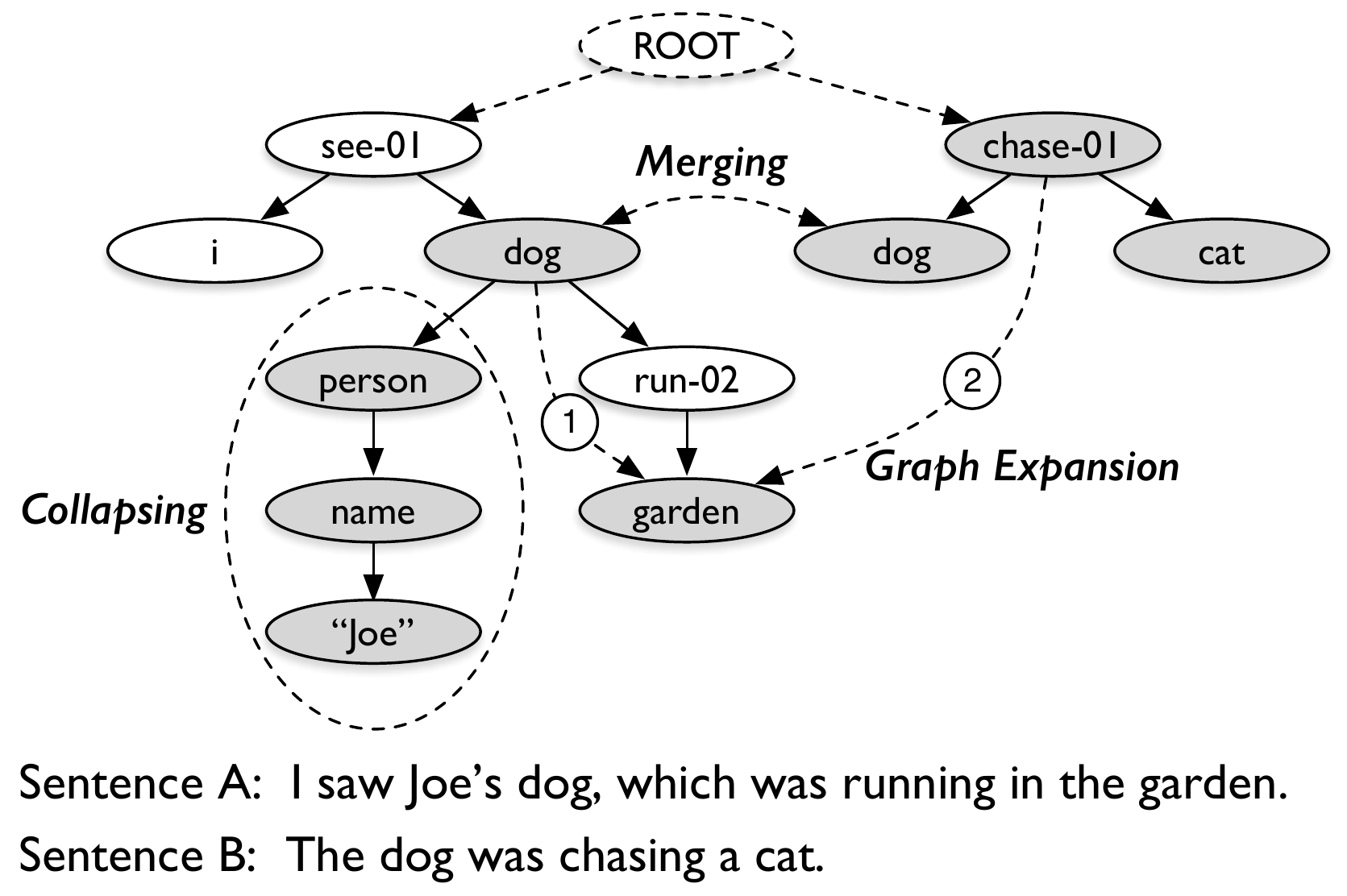}
\caption{A source graph formed from two sentence AMR graphs. 
Concept collapsing, merging, and graph expansion are demonstrated.
Edges are unlabeled.
A ``ROOT'' node is added to ensure connectivity.
(1) and (2) are among edges added through the optional expansion step, corresponding to sentence- and document-level expansion, respectively.
Concept nodes included in the summary graph are shaded.
}
\label{fig:source_graph}
\end{center}
\end{figure}

To ensure that the source graph is connected,
we add a new ``ROOT'' node and connect it to every concept that was
originally the root of a sentence graph (see Fig.~\ref{fig:source_graph}).
When we apply this procedure to the documents in our dataset (\S\ref{sec:data}),
source graphs contain 144 nodes and 221 edges on average.

We investigated  how well these automatically constructed source
graphs
cover the gold-standard summary graphs produced by AMR annotators.
Ideally, a source graph should cover all of the
gold-standard edges, so that summarization can be accomplished by
selecting a subgraph of the source graph (\S\ref{subsec:subgraphpred}).
In Table~\ref{tab:coverage}, columns one and two report labeled and unlabeled edge coverage.
`Unlabeled' counts edges as matching if both the source and destination concepts have identical labels, but ignores the edge label.

\begin{table}[t]
\setlength{\tabcolsep}{6pt}
\small
\renewcommand{\arraystretch}{1.1}
\begin{center}
\begin{tabular}{|c|rr|rr|}
\hline
 & \multicolumn{4}{c|}{\textbf{Summary Edge Coverage (\%)}} \\
&       &           & \multicolumn{2}{c|}{Expand} \\
& Labeled & Unlabeled & Sent. & Doc. \\
\hline
\hline
Train & 64.8 & 67.0 & 75.5 & 84.6 \\
Dev. & 77.3 & 78.6 & 85.4 & 91.8 \\
Test & 63.0 & 64.7 & 75.0 & 83.3 \\
\hline
\end{tabular}
\caption{Percentage of summary edges
 that can be covered by an automatically constructed source
  graph.
\label{tab:coverage}}
\end{center}
\end{table}

In order to improve edge coverage, we explore expanding the source graph by adding
every possible edge between every pair of concepts within the same sentence.
We also explored adding every possible edge between every pair of concepts in the entire source graph.
Edges that are newly introduced during expansion receive a default label `null'.
We report unlabeled edge coverage in Table~\ref{tab:coverage}, columns three and four, respectively.
Subgraph prediction became infeasable with the document-level expansion, so we conducted our experiments using only sentence-level expansion.
Sentence-level graph expansion increases the average number of edges by a factor of 15, to 3,292.
Fig.~\ref{fig:source_graph} illustrates the motivation. Document-level expansion covers the gold-standard summary edge ``chase-01'' $\rightarrow$ ``garden,'' yet the expansion is computationally prohibitive; sentence-level expansion adds an edge ``dog'' $\rightarrow$ ``garden,'' which enables the prediction of a structure with similar semantic meaning: ``Joe's dog was in the garden chasing a cat.''

\subsection{Subgraph Prediction}
\label{subsec:subgraphpred}

We pose the selection of a summary subgraph from the source graph as a
structured prediction problem that trades off among
including important information without altering its meaning, maintaining brevity, and producing fluent language~\cite{Nenkova:2011}.
We incorporate these concerns in the form of features and constraints
in the statistical model for subgraph selection.

Let $G = (V,E)$ denote the merged source graph, where each node $v \in
V$ represents a unique concept and each directed edge $e \in E$ connects two concepts.
$G$ is a connected, directed, node-labeled graph.
Edges in this graph are unlabeled, and edge labels are not predicted
during subgraph selection. 
We seek to maximize a score that factorizes over graph nodes and
edges that are included in the summary graph.  For subgraph $(V', E')$:
\begin{equation}
\label{eqn:obj}
\mathit{score}(V', E'; \boldsymbol{\theta},\boldsymbol{\psi}) = \displaystyle \sum_{v \in V'} {{\boldsymbol\theta ^{\top}}\mathbf{f}(v)} + \sum_{e \in {E'}} {{\boldsymbol\psi ^{\top}}\mathbf{g}(e)}
\end{equation}
where $\mathbf{f}(v)$ and $\mathbf{g}(e)$ are the feature
representations of node $v$ and edge $e$, respectively.
We describe node and edge features in Table~\ref{tab:features}.
$\boldsymbol\theta$ and $\boldsymbol{\psi}$ are vectors of empirically estimated
coefficients in a linear model.


We next formulate the selection of the subgraph using integer linear
programming (ILP; \S\ref{sec:decoding}) and describe supervised
learning
for the parameters (coefficients) from a collection of source graphs paired  with
summary graphs (\S\ref{sec:learning}).

\begin{table*}[t]
\begin{scriptsize}
\begin{center}
\setlength{\tabcolsep}{5pt}
\renewcommand{\arraystretch}{1.1}
\begin{tabular}{|l|l|l|}
\hline
Node & Concept & Identity feature for concept label\\
Features & Freq & Concept freq in the input sentence set; one binary feature defined for each frequency threshold $\tau=0/1/2/5/10$\\
& Depth & Average and smallest depth of node to the root of the sentence graph\smtcomment{source?}\jmf{fixed?}\flcomment{changed to sentence graph;}; binarized using 5 depth thresholds \smtcomment{which?}\flcomment{thresholds omitted due to length limitation..}\\
& Position & Average and foremost position of sentences containing the concept; binarized using 5 position thresholds \smtcomment{which?} \\
& Span & Average and longest word span of concept; binarized using 5 length thresholds\smtcomment{which?}; word spans obtained from JAMR  \\
& Entity & Two binary features indicating whether the concept is a named entity/date entity or not\\
& Bias & Bias term, 1 for any node\\
\hline
\hline
Edge & Label & First and second most frequent edge labels between concepts; relative freq of each label, binarized by 3 thresholds \smtcomment{which?}\\
Features & Freq & Edge frequency (w/o label, non-expanded edges) in the document sentences; binarized using 5 frequency thresholds \smtcomment{which?}\\
& Position & Average and foremost position of sentences containing the edge (without label); binarized using 5 position thresholds \smtcomment{which?}\\
& Nodes & Node features extracted from the source and target nodes (all above node features except the bias term)\\
& IsExpanded & A binary feature indicating the edge is due to graph expansion or not; edge freq (w/o label, all occurrences)\\
& Bias & Bias term, 1 for any edge\\
\hline
\end{tabular}
\caption{Node and edge features (all binarized).
}
\label{tab:features}
\end{center}
\end{scriptsize}
\end{table*}

\subsubsection{Decoding}
\label{sec:decoding}

We cast decoding as an ILP whose constraints ensure that the output
forms a connected subcomponent of the source graph.
We index source graph concept nodes by $i$ and $j$, giving the ``ROOT'' node index $0$.
Let $N$ be the number of nodes in the graph.
Let $v_i$ and $e_{i,j}$ be binary variables. $v_i$ is 1 iff source
node $i$ is included; $e_{i,j}$ is 1 iff the directed
edge from node $i$ to node $j$ is
included. 

The ILP objective to be maximized is Equation~\ref{eqn:obj}, rewritten here in the
present notation:
\begin{equation}
\label{eqn:obj2}
\displaystyle \sum_{i=1}^N v_i \underbrace{{\boldsymbol\theta
    ^{\top}}\mathbf{f}(i)}_{\mbox{node score}} + \sum_{(i,j) \in E}
e_{i,j} \underbrace{{\boldsymbol\psi
    ^{\top}}\mathbf{g}(i,j)}_{\mbox{edge score}}
\end{equation}
Note that this objective is linear in $\{v_i, e_{i,j}\}_{i,j}$ and
that features and coefficients can be folded into node and edge scores
and treated as constants during decoding.

Constraints are required to ensure that the selected nodes and edges
form a valid graph.
In particular, if an edge $(i,j)$ is selected ($e_{i,j}$ takes value of
1), then both its endpoints $i$, $j$ must be included:
\begin{equation}
{v_i} - {e_{i,j}} \ge 0, \quad {v_j} - {e_{i,j}} \ge 0, \quad \forall
i,j \le N 
\label{eqn:constr_node}
\end{equation}

Connectivity is enforced using a set of single-commodity flow variables $f_{i,j}$, each taking a non-negative integral value, representing the flow from node $i$ to $j$. 
The root node sends out up to $N$ units of flow, one to reach each
included node (Equation~\ref{eqn:constr_flow_start}).
Each included node consumes one unit of flow, reflected as the
difference between incoming and outgoing flow (Equation~\ref{eqn:constr_flow_consume}).
Flow may only be sent over an edge if the edge is included (Equation~\ref{eqn:constr_flow_edge}).
\begin{eqnarray}
\label{eqn:constr_flow_start}
\sum_i {f_{0,i}} - \sum_i {v_i} = 0, \\
\label{eqn:constr_flow_consume}
\sum_{i} {{f_{i,j}} - } \sum_{k} {{f_{j,k}} - v_j = 0},
\quad\forall j \le N, \\
\label{eqn:constr_flow_edge}
N \cdot {e_{i,j}} - {f_{i,j}} \ge 0, \quad \forall i,j \le N.
\end{eqnarray}

The AMR representation allows graph reentrancies (concept nodes having multiple parents),
yet reentrancies are rare; about 5\% of edges are re-entrancies in our dataset.
In this preliminary study we force the summary graph to be tree-structured, requiring that there is at most one incoming edge for each node:

\begin{equation}
\label{eqn:constr_edge}
\sum_{j} {e_{i,j} \le 1},  \quad \forall j \le N.
\end{equation}

Interestingly, the formulation  so far
equates to an ILP for solving the prize-collecting Steiner tree
problem (PCST; Segev, 1987)\nocite{segev-87}, which is known to be NP-complete
\cite{karp-72}.
Our ILP formulation is modified from that of~\newcite{ljubic-06}.
Flow-based constraints for tree structures have also previously been used in NLP for dependency parsing~\cite{Martins:2009:ACL} and sentence compression~\cite{Thadani:2013}. 
In our experiments, we use an exact
ILP solver,\footnote{\url{http://www.gurobi.com}} though many approximate methods are available.

Finally, an optional constraint can be used to fix the size of the summary
graph (measured by the number of edges) to $L$:
\begin{equation}
\label{eqn:constr_len}\sum_{i} \sum_{j} {e_{i,j} = L}
\end{equation}
The performance of summarization systems depends strongly on their compression rate,
so systems are only directly comparable when their compression rates are similar \cite{napoles-11}.
$L$ is supplied to the system to control summary graph size.
\flcomment{Using number of concepts as threshold could be better. We couldn't produce summaries of desired length (typically measured by number of words) in either case due to the text generation step.}

\subsubsection{Parameter Estimation}
\label{sec:learning}

Given a collection of input and output pairs (here, source graphs and
summary graphs), a natural starting place for learning the coefficients
$\boldsymbol{\theta}$ and $\boldsymbol{\psi}$ is the structured
perceptron \cite{Collins:2002}, which is easy to implement and often
performs well.  Alternatively, incorporating factored cost functions through a
structured hinge loss leads to a structured support vector machine
(SVM; Taskar et al., 2004) \nocite{taskar-04} which can be learned
with a very similar stochastic optimization algorithm.
In our scenario, however, the gold-standard summary
graph may not actually be a subset of the source graph.
In machine translation, \emph{ramp loss} has been found to work well in situations where the gold-standard output may not even be in the hypothesis space of the model 
\cite{Gimpel:2012}.
The structured perceptron, hinge, and ramp losses are compared in Table~\ref{tab:losses}.

We explore learning by minimizing each of the perceptron, hinge, and ramp losses, each optimized using
Adagrad~\cite{Duchi:2011},
a stochastic optimization
procedure.  Let $\beta$ be one model parameter (coefficient from
$\boldsymbol{\theta}$ or $\boldsymbol{\psi}$).
Let $g^{(t)}$ be the subgradient of
the loss on the instance considered on the $t^{th}$ iteration with
respect to $\beta$. 
Given an initial step size $\eta$, the update for $\beta$ on iteration $t$ is:
\begin{equation}
\beta^{(t+1)} \leftarrow \beta^{(t)} - \frac{\eta}{\sqrt{\sum_{\tau=1}^t
    (g^{(\tau)})^2}} g^{(t)}
\end{equation}

\begin{table*}
\centering \begin{tabular}{lrcl}
Structured perceptron loss: & $-\mathit{score}(G^\ast)$ & $+$ &
$\displaystyle \max_{G} \mathit{score}(G)$ \\
Structured hinge loss: & $-\mathit{score}(G^\ast)$ & $+$ &
$\displaystyle \max_{G} (\mathit{score}(G) + \mathit{cost}(G; G^\ast))$
\\
Structured ramp loss: & $-\displaystyle \max_{G} (\mathit{score}(G) - \mathit{cost}(G; G^\ast))$ & $+$ &
$\displaystyle \max_{G} (\mathit{score}(G) + \mathit{cost}(G; G^\ast))$
\end{tabular}
\caption{Loss functions minimized in parameter estimation.
$G^\ast$ denotes the gold-standard summary graph.
$\mathit{score}(\cdot)$ is as defined in Equation~\ref{eqn:obj}. 
$\mathit{cost}(G; G^\ast)$ penalizes
each vertex or edge in $G \cup G^\ast \setminus (G \cap G^\ast)$.
Since $\mathit{cost}$ factors just like the scoring function,
each $\max$ operation can be accomplished using a variant of ILP
decoding (\S\ref{sec:decoding}) in which the cost is incorporated into the linear objective while the constraints remain the same.
\label{tab:losses}}
\vspace{0.05in}
\end{table*}

\section{Generation}
\label{sec:nlg}


Generation from AMR-like representations has received some 
attention, e.g., by \newcite{Langkilde:1998} who described a
statistical method.  Though we know of work in progress driven by
the goal of machine translation using AMR, there is currently no
system available.

We therefore use a heuristic approach to generate a bag of words.
Given a predicted subgraph, a system summary is created by finding the
most frequently aligned word span for each concept node.  (Recall that
the JAMR parser provides these alignments; \S\ref{sec:amr}).
The words in the resulting spans are generated in no particular order. 
While this is not a natural language summary, it is suitable for
unigram-based summarization evaluation methods like ROUGE-1.




\section{Experiments}
\label{sec:experiments}

\begin{table*}[th]
\setlength{\tabcolsep}{11pt}
\renewcommand{\arraystretch}{1.1}
\begin{center}
\begin{footnotesize}
\begin{tabular}{|l|l|rrr|r|rrr|}
\cline{3-9}
\multicolumn{2}{c|}{} & \multicolumn{4}{c|}{\textbf{Subgraph Prediction}} & \multicolumn{3}{c|}{\textbf{Summarization}}\\
\multicolumn{2}{c|}{} & \multicolumn{3}{c}{\textbf{Nodes}} & \multicolumn{1}{c|}{\textbf{Edges}} & \multicolumn{3}{c|}{\textbf{ROUGE-1}}\\
\multicolumn{2}{c|}{} & P (\%) & R (\%) & \multicolumn{1}{r}{F (\%)} & F (\%) & P (\%) & R (\%) & F (\%)\\
\cline{3-9}
\hline
\textbf{gold-} & Perceptron & 39.6 & 46.1 & 42.6 & 24.7 & 41.4 & 27.1 & 32.3 \\
\textbf{standard} & Hinge & 41.2 & 47.9 & 44.2 & 26.4 & 42.6 & 28.3 & 33.5 \\
\textbf{parses}  & Ramp & \textbf{54.7} & \textbf{63.5} & \textbf{58.7} & \textbf{39.0} & \textbf{51.9} & \textbf{39.0} & \textbf{44.3} \\
 & Ramp + Expand & 53.0 & 61.3 & 56.8 & 36.1 & 50.4 & 37.4 & 42.8\\
\cline{2-9}
 & Oracle & {75.8} & {86.4} & {80.7} & {52.2} & \multirow{2}{*}{89.1} & \multirow{2}{*}{52.8} & \multirow{2}{*}{{65.8}} \\
  & Oracle + Expand & 78.9 & {90.1} & {83.9} & {64.0} & &  & \\
\hline
\hline
\textbf{JAMR} & Perceptron & 42.2 & 48.9 & 45.2 & 14.5 & 46.1 & 35.0 & 39.5 \\
\textbf{parses} & Hinge & 41.7 & 48.3 & 44.7 & 15.8 & 44.9 & 33.6 & 38.2 \\
 & Ramp & \textbf{48.1} & \textbf{55.6} & \textbf{51.5} & \textbf{20.0} & {50.6} & {40.0} & {44.4} \\
 & Ramp + Expand & 47.5 & 54.6 & 50.7 & 19.0 & \textbf{51.2} & \textbf{40.0} & \textbf{44.7}\\
\cline{2-9}
 & Oracle & {64.1} & {74.8} & {68.9} & {31.1} &  \multirow{2}{*}{87.5} &  \multirow{2}{*}{43.7} &  \multirow{2}{*}{{57.8}} \\
 & Oracle + Expand & {66.9} & {76.4} & {71.2} & {46.7} &
 && \\
\hline
\end{tabular}
\caption{Subgraph prediction and summarization (to bag of words)
  results on test set. Gold-standard AMR annotations are used for model training in all conditions. ``+ Expand'' means the result is obtained using source graph with expansion; 
edge performance is measured ignoring labels.
\smtcomment{also, the bolding in this table is weird. Why are oracle
  results bolded? Why is the ``Ramp'' row in ROUGE-1 bolded, even
  though ``Ramp+Expand'' is better?}
\nascomment{agree with Sam's comments.  should always explain the use
  of boldface in the caption. don't boldface oracles.}
\flcomment{changed.}
}
\label{tab:results}
\end{footnotesize}
\end{center}
\end{table*}


In Table~\ref{tab:results}, we report the  performance of
subgraph prediction and end-to-end summarization  on the test set,
using gold-standard and automatic AMR parses for the input.
Gold-standard AMR annotations are used for model training in all conditions.
During testing, we apply the trained model to source graphs constructed using either gold-standard or JAMR parses.
In all of these experiments, we use the number of edges in the gold-standard summary graph
to fix the number of edges in the predicted subgraph, allowing
direct comparison across conditions. 

Subgraph prediction is evaluated against the gold-standard AMR graphs
on summaries.
We report precision, recall, and $F_1$ for nodes, and
$F_1$ for edges.\footnote{Precision, recall, and $F_1$ are equal since
the number of edges is fixed.}




Oracle results for the subgraph prediction stage are obtained using the ILP decoder to
minimize the cost of the output graph, given the gold-standard.
We assign wrong nodes and edges a score of $-1$, correct nodes and edges a score of $0$, then decode with the same structural constraints as in subgraph prediction.
The resulting graph is the best summary graph in the hypothesis space of our model, and provides an upper bound on performance achievable within our framework.
Oracle performance on node prediction is in the range of 80\% when using gold-standard AMR annotations, and 70\% when using JAMR output.
Edge prediction has lower performance, yielding 52.2\% for gold-standard and 31.1\% for JAMR parses.
When graph expansion was applied, the numbers increased to 64\% and 46.7\%, respectively.
The uncovered summary edge (i.e., those not covered by source graph) is a major source for low recall values on edge prediction (see Table~\ref{tab:coverage}); graph expansion slightly alleviates this issue.

Summarization is evaluated by comparing system summaries against reference summaries, using ROUGE-1 scores~\cite{Lin:2004}\footnote{ROUGE version 1.5.5 with options `-e data -n 4 -m -2 4 -u -c 95 -r 1000 -f A -p 0.5 -t 0 -a -x'}.
System summaries are generated using the heuristic approach presented in \S\ref{sec:nlg}:
given a predicted subgraph, the approach finds the most frequently aligned word span for each concept node, and then puts them together as a bag of words.
ROUGE-1 is particularly usefully for evaluating such less well-formed summaries, such as those generated from speech transcripts~\cite{Liu:2013}.

Oracle summaries are produced by taking the gold-standard AMR parses of the reference summary, obtaining the most frequently aligned word span for each unique concept node using the JAMR aligner (\S\ref{sec:amr}), and then generating a bag of words summary. 
Evaluation of oracle summaries is performed in the same manner as for system summaries.
The above process does not involve graph expansion, so summarization performance is the same for the two conditions ``Oracle'' and ``Oracle + Expand.''

We find that JAMR parses are a large source of degradation of edge prediction performance, and a smaller but still significant source of degradation for concept prediction. 
Surprisingly, using JAMR parses leads to slightly improved ROUGE-1 scores.
Keep in mind, though, that under our bag-of-words generator, ROUGE-1 scores only depend on concept prediction and are unaffected by edge prediction.\jmf{is this true? which version of ROUGE do we use, i.e., does it look at function words?}
\flcomment{yep, since we read word spans off the predicted concepts instead of using a text generator. I added ROUGE version and options in earlier text.}
The oracle summarization results, 65.8\% and 57.8\% $F_1$ scores for gold-standard and JAMR parses, respectively, further suggest that improved graph summarization models (step 2) might benefit from future improvements in AMR parsing (step 1).

Across all conditions and both evaluations, we find that incorporating a cost-aware loss function (hinge vs.~perceptron) has little effect, but that using ramp loss
leads to substantial gains.





In Table~\ref{tab:results}, we show detailed results with and without graph expansion. 
``+ Expand'' means the results are obtained using the expanded source graph.
We find that graph expansion only marginally affects system performance.
Graph expansion slightly hurts the system performance on edge prediction. 
For example, using ramp loss with JAMR parser as input, we obtained 50.7\% and 19.0\% for node and edge prediction with graph expansion; 51.5\% and 20.0\% without edge expansion.
On the other hand, it increases the oracle performance by a large margin.
This suggests that with more training data, or a more sophisticated model that is able to better discriminate among the enlarged output space,
graph expansion still has promise to be helpful.


\section{Related and Future Work}
\label{sec:related}



According to ~\newcite{Dang:2008}, the majority of competitive summarization
systems are extractive, selecting representative sentences from input
documents and concatenating them to form a summary.   This is often
combined with sentence compression, allowing more sentences to be
included within a budget.
ILPs and approximations have been used to
encode compression and extraction \cite{McDonald:2007,Martins:2009,Gillick:2009:NAACL,Kirkpatrick:2011,Almeida:2013,Chen:2014}. 
Other decoding approaches have included a greedy method exploiting submodularity~\cite{Lin:2010:NAACL}, document reconstruction~\cite{He:2012}, and graph cuts~\cite{Qian:2013}, among others.



Previous work on abstractive summarization has explored 
user studies that compare extractive with NLG-based abstractive
summarization~\cite{Carenini:2008}.
~\newcite{Ganesan:2010} propose to construct summary sentences
by\nascomment{this is not well-formed, and I don't understand it:  }\flcomment{slightly reworded.} repeatedly searching the highest scored graph paths. 
~\cite{Gerani:2014} generate abstractive summaries by modifying discourse parse trees.
Our work is similar in spirit to \newcite{Cheung:2014}, which splices and recombines dependency parse trees to produce abstractive summaries.
In contrast, our work operates on semantic graphs, taking advantage of the recently developed AMR Bank.

  
Also related to our work are graph-based summarization methods
\cite{Vanderwende:2004,Erkan:2004,Mihalcea:2004}.\smtcomment{our method is ``graph-based'' also, but i think you mean it in a different sense here. need to use a different term, or explain what we mean by ``graph-based'' here.}\flcomment{clarified a bit.}
~\newcite{Vanderwende:2004} transform input to logical forms, score nodes using PageRank, and grow the graph from high-value nodes using heuristics. 
In~\newcite{Erkan:2004} and \newcite{Mihalcea:2004}, the graph connects
surface terms that co-occur.
In both cases, the graphs are constructed based on surface text; it is not a representation of propositional semantics like AMR.  However, future work might explore similar
graph-based calculations to contribute features for subgraph selection
in our framework.


Our constructed source graph can easily reach ten times or more of the size of a sentence dependency graph.
Thus more efficient graph decoding algorithms, e.g., based on Lagrangian relaxation or approximate algorithms, may be explored in future work.
Other future directions may include jointly performing subgraph and edge label prediction; exploring a full-fledged pipeline that consists of an automatic AMR parser, a graph-to-graph summarizer, and a AMR-to-text generator; and devising an evaluation metric that is better suited to abstractive summarization.

Many domains stand to eventually benefit from summarization.
These include books, audio/video segments, and legal texts.

\section{Conclusion}
\label{sec:conclusion}

We have introduced a statistical abstractive summarization framework driven by
the Abstract Meaning Representation.  The centerpiece of the approach
is a structured prediction algorithm that transforms semantic graphs
of the input into a single summary semantic graph.  Experiments show
the approach to be promising and suggest directions for future research.

\section*{Acknowledgments}
The authors thank three anonymous reviewers for their insightful input. 
We are grateful to Nathan Schneider, Kevin Gimpel, Sasha Rush, and the ARK group for valuable discussions.
The research was supported by NSF grant SaTC-1330596, DARPA grant
FA8750-12-2-0342 funded under the DEFT program, the U. S. Army
Research Laboratory and the U. S. Army Research Office under
contract/grant number W911NF-10-1-0533, and by IARPA via 
DoI/NBC contract number D12PC00337. The views and conclusions contained
herein are those of the authors and should not be
interpreted as necessarily representing the official
policies or endorsements, either expressed or implied, of the
sponsors.

\bibliographystyle{naaclhlt2015}
\bibliography{semsumm}

\end{document}